\documentclass[sigconf]{acmart}
\AtBeginDocument{%
  }

\copyrightyear{2026}
\acmYear{2026}
\setcopyright{cc}
\setcctype{by-nc-nd}
\acmConference[KDD '26]{Proceedings of the 32nd ACM SIGKDD Conference on Knowledge Discovery and Data Mining V.2}{August 09--13, 2026}{Jeju Island, Republic of Korea}
\acmBooktitle{Proceedings of the 32nd ACM SIGKDD Conference on Knowledge Discovery and Data Mining V.2 (KDD '26), August 09--13, 2026, Jeju Island, Republic of Korea}
\acmDOI{10.1145/3770855.3818867}
\acmISBN{979-8-4007-2259-2/2026/08}

\usepackage[ruled,vlined]{algorithm2e}

\usepackage{hyperref}
\usepackage{makecell}
\usepackage{multirow}
\usepackage{booktabs} 
\usepackage{mathrsfs}
\usepackage{cleveref}

\newtheorem{theorem}{Theorem}[section]
\newtheorem{proposition}[theorem]{Proposition}

\crefname{equation}{Eq.}{Eqs.}
\Crefname{equation}{Equation}{Equations}
\crefname{figure}{Fig.}{Figs.}
\Crefname{figure}{Figure}{Figures}
\crefname{table}{Tab.}{Tabs.}
\Crefname{table}{Table}{Tables}
\crefname{algocf}{Alg.}{Algs.}
\Crefname{algocf}{Algorithm}{Algorithms}
\crefname{section}{Sec.}{Secs.}
\Crefname{section}{Section}{Sections}
\crefname{appendix}{App.}{Apps.}
\Crefname{appendix}{Appendix}{Appendices}

\crefname{theorem}{Thm.}{Thms.}
\Crefname{theorem}{Theorem}{Theorems}
\crefname{lemma}{Lem.}{Lems.}
\Crefname{lemma}{Lemma}{Lemmas}
\crefname{definition}{Def.}{Defs.}
\Crefname{definition}{Definition}{Definitions}
\crefname{corollary}{Cor.}{Cors.}
\Crefname{corollary}{Corollary}{Corollaries}
\crefname{remark}{Rem.}{Rems.}
\Crefname{remark}{Remark}{Remarks}
\crefname{proposition}{Prop.}{Props.}
\Crefname{proposition}{Proposition}{Propositions}
\crefname{proof}{Pr.}{Prs.}
\Crefname{proof}{Proof}{Proofs}

\begin{document}

\title{Towards Robust EEG Decoding Based on Riemannian Self-Attention}

\author{Shaocheng Jin}
\affiliation{%
  \institution{School of Artificial Intelligence and Computer Science}
  \institution{Jiangnan University}
  \city{Wuxi}
  \country{China}}
\email{shaochengjin.ai@gmail.com}

\author{Tao Zhou}
\affiliation{%
  \institution{School of Artificial Intelligence and Computer Science}
  \institution{Jiangnan University}
  \city{Wuxi}
  \country{China}}
\email{taozhou@jiangnan.edu.cn}

\author{Rui Wang}
\authornote{Rui Wang is the corresponding author.}
\affiliation{%
  \institution{School of Artificial Intelligence and Computer Science}
  \institution{Jiangnan University}
  \city{Wuxi}
  \country{China}}
\email{cs_wr@jiangnan.edu.cn}

\author{Ziheng Chen}
\affiliation{%
  \institution{Department of Information Engineering and Computer Science}
  \institution{University of Trento}
  \city{Trento}
  \country{Italy}}
\email{ziheng_ch@163.com}

\author{Xiaoqing Luo}
\affiliation{%
  \institution{School of Artificial Intelligence and Computer Science}
  \institution{Jiangnan University}
  \city{Wuxi}
  \country{China}}
\email{xqluo@jiangnan.edu.cn}

\author{Xiao-Jun Wu}
\affiliation{%
  \institution{School of Artificial Intelligence and Computer Science}
  \institution{Jiangnan University}
  \city{Wuxi}
  \country{China}}
\email{wu_xiaojun@jiangnan.edu.cn}

\author{Josef Kittler}
\affiliation{%
  \institution{Centre for Vision, Speech and Signal Processing (CVSSP)}
  \institution{University of Surrey}
  \city{Guildford}
  \country{UK}}
\email{j.kittler@surrey.ac.uk}

\renewcommand{\shortauthors}{Shaocheng Jin et al.}

\begin{abstract}
    Brain–Computer Interface (BCI) based on electroencephalography (EEG) enables direct interaction between the brain and external environments and has significant applications in assistive technologies, medical rehabilitation, and entertainment. Recently, EEG decoding methods based on Symmetric Positive Definite (SPD) learning have demonstrated superior performance. However, these methods typically employ basic network architectures and do not explicitly capture local relationships between EEG signals. This limitation is problematic for EEG signals due to their inherently low Signal-to-Noise Ratio (SNR). Moreover, most existing Riemannian manifold-based methods are restricted to specific metrics. The most widely used is the Affine-Invariant Metric (AIM). However, it has a quadratic dependency on the SPD matrices and can not handle ill-conditioned SPD matrices (ICSM), which hinders the effectiveness of networks. In contrast, the Bures-Wasserstein Metric (BWM)  exhibits linear dependence on SPD matrices and demonstrates superior performance for ill conditioning. To overcome these challenges, we propose a Riemannian self-attention network based on the BWM. Additionally, the recently introduced power-deformed generalized BWM ($\theta$-GBWM) parameterizes the vanilla BWM via an SPD matrix and matrix power deformation. This metric provides a more nuanced representation of the geometric structure of the SPD manifold. Consequently, we extend our model to a learnable version. For simplicity, we refer to it as GBWAtt. Experimental results on three EEG benchmarking datasets validate the robustness and effectiveness of our proposed method. The code is available at \url{https://github.com/jjscc/GBWAtt}.
\end{abstract}



\maketitle

\section{Introduction}
\label{sec:introduction}

In recent years, the rapid advancement of deep learning has significantly accelerated progress in EEG decoding~\cite{roy2019deep,bore2021long}, primarily owing to its ability to extract expressive spatiotemporal representations through nonlinear, end-to-end representation learning. Among these methods, Convolutional Neural Networks (ConvNets) are widely adopted and have demonstrated outstanding performance in a variety of tasks, such as image recognition and object detection~\cite{lecun2002gradient,resnet,howard2017mobilenets,huang2017densely}. For EEG decoding, ConvNets can provide greater flexibility than traditional spatial and temporal filtering methods, enabling the models to automatically optimize transformations of EEG signals through data-driven training~\cite{matt17,matt18,matt19}. Beyond the rapid development of deep learning-based EEG decoders, geometric learning approaches, particularly those based on Riemannian geometry, have also been introduced into the field of BCI, achieving promising results~\cite{lotte2007review,yger2016riemannian,congedo2017riemannian}.

While these advantages have expanded the algorithmic 
landscape for EEG decoding, Motor Imagery (MI) based EEG classification remains a central and challenging task in BCI research, mainly due to the substantial inter-subject variability and overlapping of the neural patterns associated with different MI tasks. The dominant perspective of MI-based EEG classification lies in temporal-domain analysis, where most approaches leverage the second-order statistics of EEG signals, particularly the spatial covariance matrix. This matrix can capture the linear relationships between EEG channels by estimating the joint variability of their time-series signals. Its diagonal elements correspond to the variance of individual channels, reflecting the power of neural oscillations at specific electrode sites, whereas the off-diagonal elements encode inter-channel covariances, providing a measure of spatial correlation patterns~\cite{yger2013review,barachant2010riemannian}. Such characteristics have proven highly effective in characterizing and analyzing multichannel time-series data for BCI~\cite{bci}.

However, the space formed by a set of non-singular covariance matrices is not a vector space. Instead, it is a Riemannian manifold, \textit{i.e.}, SPD manifold~\cite{tang2020generalized}. This intrinsic non-Euclidean structure hinders the straightforward application of Euclidean learning methods to such data. To narrow this gap, several Riemannian metrics have been developed, such as the Log-Euclidean Metric (LEM)~\cite{ArsV} and the AIM~\cite{PenX}, laying the groundwork for the adaptation of Euclidean computations to the context of SPD manifolds. Based on these Riemannian geometries, a series of effective learning algorithms have been established for SPD matrix learning. Some strategies map the manifold-valued data into a flat Euclidean space by tangent approximation~\cite{mcc,stcd}, while others employ Riemannian kernel functions~\cite{cdl,vem,rcdl}, to embed them into a Reproducing Kernel Hilbert Space (RKHS). Nevertheless, the aforementioned approaches might yield suboptimal results due to the potential distortion of the geometrical structure of the input data during approximation or embedding. To tackle this issue, several geometry-aware methods~\cite{leml,cml,spdml,ps3d} directly build an embedding mapping between manifolds, ensuring effective preservation of the original Riemannian geometry. Despite these advancements, the shallow learning mechanism of this type of approach restricts its ability to generate more informative and task-discriminative  representations. 

As a countermeasure, considerable efforts have been dedicated to extending the Euclidean Deep Learning (EucDL) paradigm to the context of Riemannian manifolds. Among these, SPD matrix network (SPDNet)~\cite{spdnet} stands out as a foundational Riemannian neural network, offering a multi-stage, end-to-end nonlinear learning mechanism specifically designed for SPD matrices. In addition, many other EucDL architectures have been successfully generalized to the Riemannian manifolds, such as Recurrent Neural Network (RNN)~\cite{chakraborty2018statistical,nguyen2022gyro}, Graph Convolutional Networks (GCN)~\cite{hgrnet,ju2023graph}, ConvNets~\cite{manifoldnet,chen2023riemannian}, and Transformers~\cite{pan2022matt}. Besides, some researchers have studied the basic building blocks for Riemannian networks, such as Riemannian pooling~\cite{symnet,chen2025understanding}, Riemannian classifier~\cite{spdmlr,nguyen2023building}, and Riemannian batch normalization~\cite{spdnetbn,wang2025learning}.

Although SPD neural networks can capture non-Euclidean correlations in multichannel EEG signals, most existing models use relatively simple architectures and insufficiently model local relationships among EEG-derived SPD features, which is particularly problematic under the low-SNR nature of EEG signals. This motivates us to explore a Riemannian self-attention (RieAtt) mechanism on the SPD manifold. A key challenge in SPD representation learning is the presence of ill-conditioned SPD matrices (ICSM). Since covariance matrices are positive semi-definite, a common regularization strategy is: $X_i \gets X_i+\lambda I_n$, where $I_n$ is an $n$-by-$n$ identity matrix and $\lambda$ is a perturbation parameter. Although this strategy enforces positive definiteness, $\lambda$ must be small to avoid distorting the eigenvalue space~\cite{cdl,spdml,spdnet,spdnetbn}. As a result, the regularized SPD matrices may still have large condition numbers.

\begin{table}[t] 
\renewcommand\arraystretch{1.3}
\centering
\caption{Comparison of Riemannian Metric between AIM and BWM.}
\label{comparer_sum}
\resizebox{0.9\linewidth}{!}{
\begin{tabular}{l|c}
\toprule
 Metric &  ${g}_{X}(S_{1},S_{2})$ \\
\midrule
AIM  &  $\mathrm{tr}(X^{-1}S_{1}X^{-1}S_{2})=\mathrm{vec}(S_{1})^{\top}(X\otimes X)^{-1}\mathrm{vec}(S_{2})$ \\
\midrule
BWM  &  $\frac{1}{2}{\mathrm{tr}}(\mathcal{L}_{X}(S_{1})S_{2})=\frac{1}{2}\mathrm{vec}(S_{1})^{T}(X \oplus X)^{-1}\mathrm{vec}(S_{2})$ \\
\bottomrule
\end{tabular}}
\end{table}

Among the Riemannian metrics used in SPD neural networks, AIM is the most widely used one. However, AIM is not numerically unstable in SPD matrix computation, ~\cite{han2021riemannian}. In contrast, the recently proposed BWM exhibits good numerical stability. As shown in \cref{comparer_sum}, AIM exhibits a quadratic dependence on the SPD matrices, but the BWM shows a linear dependence, rendering it more numerically stable for learning SPD matrices through the Kronecker sum $\oplus$, especially in the ill-conditioned scenario~\cite{han2021riemannian}. Therefore, we design a Riemannian self-attention mechanism on the SPD manifold based on the BW geometry.

Besides, Han et al.~\cite{han2021generalized} updated BWM into a generalized version, named  GBWM, with an SPD parameter. This metric connects the BWM and AIM (locally). Subsequently, Wang et al.~\cite{wang2025learning} incorporated matrix power deformation into GBWM and extended it to the power-deformed GBWM ($\theta$-GBWM), which can be viewed as a locally deformed AIM and LEM-like metric when $\theta \to 0$. Therefore, by setting the SPD parameter in $\theta$-GBWM to be learnable, our proposed self-attention model is extended into the one based on the $\theta$-GBWM. Compared with the existing AIM-based learning methods, our method can not only better deal with ICSM, but also adapt to the evolving latent SPD geometry through the dynamic geometric parameter. For simplicity, we refer to the proposed method as GBWAtt. Our main contributions are summarized as follows: 
\begin{itemize}
        \item[$\bullet$] 
        \textbf{An effective geometric deep learning framework}: We propose an SPD self-attention model based on the Bures-Wasserstein geometry.   
        \item[$\bullet$] 
        \textbf{A learnable and flexible method}: Our method is extended to a learnable version with an SPD parameter and matrix power-based nonlinear deformation of the metric geometry.
        \item[$\bullet$] 
        \textbf{Experimental validity}: Extensive empirical validations of our model on three EEG benchmarking datasets.
\end{itemize}

\section{Preliminary}
\label{sec:preliminary}

\subsection{Notions}
In this paper, the SPD manifold is composed of a set of SPD matrices, defined as $\mathrm{Sym}^{+}_n:= \{X\in \mathbb{R}^{n\times n}, X =X^{\rm{T}},v^{\top}Xv>0,\forall v\in\mathbb{R}^n\backslash\{0_n\}\}$. The tangent space at $X$ is signified as ${T}_{X}\mathrm{Sym}^{+}_n \cong \mathrm{Sym}_n $, where $\mathrm{Sym}_n$ is the Euclidean space of real symmetric matrices, denoted as $\mathrm{Sym}_n:=\{S: S \in\mathbb{R}^{n\times n}, S = S^{\top}\}$. Here, $\exp(X)$ and $\log(X)$ denote the usual matrix exponential and logarithm of $X$, respectively. For SPD, it can be written as:
\begin{equation}
    \exp(X)=V \mathrm{diag}(\exp(\sigma_1),\cdots,\exp(\sigma_n)) V^{\top},
\end{equation}
\begin{equation}
    \log(X)=V \mathrm{diag}(\log(\sigma_1),\cdots,\log(\sigma_n)) V^{\top},
\end{equation}
where  $\sigma_i\,(i\in \{1,\cdots,n\})$ represents the $i^{th}$ eigenvalue of $X$, and $V$ is the eigenvector matrix of $X$. Other notations will be introduced in appropriate paragraphs.

\subsection{Bures-Wasserstein Geometry}
\label{bwm}
A Riemannian metric $g$ is a function that is smooth, bilinear, and symmetric positive definite on the tangent space ${T}_{X}\mathrm{Sym}^{+}_n$, for any given $X\in\mathrm{Sym}^{+}_n$.
Given $X\in\mathrm{Sym}^{+}_n$, the Riemannian metric at $X$ is denoted as: $g_X:{T}_{X}\mathrm{Sym}^{+}_n\times {T}_{X}\mathrm{Sym}^{+}_n\to \mathbb{R}$, which is often written as an inner product $\langle\cdot ,\cdot\rangle_{X}$. The induced norm of a tangent vector $S\in {T}_{X}\mathrm{Sym}^{+}_n$ is given by $\left \| S \right \| _X = \sqrt{\langle S,S\rangle_{X}} $. The Riemannian metric related to the Bures-Wasserstein distance is known as the Bures-Wasserstein Metric (BWM). For an eigenvalue decomposition $X=V\Sigma V^{\top}\in \mathrm{Sym}^{+}_n$ with $\Sigma={\operatorname{diag}}(\sigma_{1},\dots,\sigma_{n})$ and $S=V\hat{S}V^{\top}\in \mathrm{Sym}_n$, the formula is
\begin{equation}
\label{inner1}
    {g}_X^{\mathrm{BW}}(S,S) = {g}_{\Sigma}^{\mathrm{BW}}(\hat{S},\hat{S}) = \frac{1}{2}\sum_{i,j}\frac{1}{\sigma_{i}+\sigma_{j}} \hat{S}_{i,j}^{2}.
\end{equation}
The Lyapunov equation $X\mathcal{L}_{X}(S)+\mathcal{L}_{X}(S)X=S, S\in\mathrm{Sym}_n $ allows us to implicitly define the linear map $\mathcal{L}_{X}$: $\mathrm{Sym}_n\to\mathrm{Sym}_n$, which can also be used to express the BWM~\cite{malago2018wasserstein}:
\begin{equation}
\label{inner2}
\begin{split}
        {g}_{X}^{\mathrm{BW}}(S_{1},S_{2})=\frac{1}{2}{\mathrm{tr}}(S_{1}\mathcal{L}_{X}(S_{2}))={\mathrm{tr}}(\mathcal{L}_{X}(S_{1})X\mathcal{L}_{X}(S_{2})),
\end{split}        
\end{equation}
where $S_{1},S_{2}\in{T}_{X}\mathrm{Sym}^{+}_n$, 
 $\mathrm{tr}(S)$ denotes the trace of $S$.
The Lyapunov operator can be solved by eigenvalue decomposition~\cite{han2021riemannian}:
\begin{equation}
\label{ly}
    \mathcal{L}_{X}({S}_{1})= V\left[\frac{{S'_{1}}_{i,j}}{{\delta}_{i}+{\delta}_{j}}\right]_{i,j}V^{\top},
\end{equation}
where $S'_{1} = V^{\top} S_{1}V$, $X = V\Delta V^{\top}$ is the eigenvalue decomposition of $X\in \mathrm{Sym}^{+}_n$. Following this, the Riemannian distance that corresponds to the BWM can be deduced as
\begin{equation}
\label{dist}
    d_{\mathrm{BW}}^{2}(X_{1},X_{2})=\mathrm{tr}(X_{1})+\mathrm{tr}(X_{2})-2\mathrm{tr}(X_{1}^{\frac{1}{2}}X_{2}X_{1}^{\frac{1}{2}})^{\frac{1}{2}}.
\end{equation}
The derivation of Eq. (\ref{dist}) is given in ~\cite{bhatia2019bures}.
\begin{proposition}
    A geodesic on a manifold, denoted as $\gamma [0,1]\longrightarrow \mathrm{Sym}^{+}_n$. From~\cite{malago2018wasserstein}, we can know that the geodesic is not complete on the Riemannian manifold $(\mathrm{Sym}^{+}_n,{g}^{\mathrm{BW}})$. The geodesic, which originates at $X \in \mathrm{Sym}^{+}_n$ and is directed by a vector $S_{1}\in {T}_{X}\mathrm{Sym}^{+}_n$ can be formulated as
    \begin{equation}
    \label{geod}
    \begin{split}
    {\gamma}_{X,S_{1}}(t)=X + tS_{1}+{t}^{2}\mathcal{L}_{X}(S_{1})X\mathcal{L}_{X}(S_{1}),{t}\in{I}.
    \end{split}
    \end{equation}
    Let ${\lambda}_{\mathrm{max}}=\mathrm{max\,eig}(\mathcal{L}_{X}(S_{1}))$ and ${\lambda}_{\mathrm{min}}=\mathrm{min\,eig}(\mathcal{L}_{X}(S_{1}))$, the definition interval of the geodesic ${\gamma}_{X,S_{1}}(t)$ is the interval $I_n$ defined by
\begin{itemize}
\item If $\lambda_{\min }<0<\lambda_{\max }$, $I=\left(-1 / \lambda_{\max },-1 / \lambda_{\min }\right)$.
\item If $0 \leqslant \lambda_{\min }$, $ I=\left(-1 / \lambda_{\max },+\infty\right)$.
\item If $\lambda_{\max } \leqslant 0$, $I=\left(-\infty,-1 / \lambda_{\min }\right)$.
\end{itemize}
\end{proposition}

Consequently, the Riemannian exponential map that projects the vector $S_{1}\in{T}_{X}\mathrm{Sym}^{+}_n$ back onto the manifold can be computed as follows:
\begin{equation}
\label{exp}
\begin{split}
    \mathrm{Exp}_{X}S_{1}=\gamma_{X,S_{1}}(1) =X+S_{1}+\mathcal{L}_{X}(S_{1})X\mathcal{L}_{X}(S_{1}).
\end{split}
\end{equation}
Likewise, the Riemannian logarithmic map, which maps $X\in\mathrm{Sym}^{+}_n$ to ${T}_{X_{0}}\mathrm{Sym}^{+}_n$ at $X_{0}\in\mathrm{Sym}^{+}_n$ is given by
\begin{equation}
\label{log}
\begin{split}
   \mathrm{Log}_{X_{0}}X=(XX_{0})^{\frac{1}{2}}+(X_{0}X)^{\frac{1}{2}}-2X_{0}.
\end{split}
\end{equation}

Recently, Han et al.~\cite{han2023learning} further updated BWM into a generalized version, named Generalized Bures-Wasserstein metric (GBWM), by an SPD parameter. For any two $S_{1}, S_{2} \in T_{X}\mathrm{Sym}^{+}_n$, the GBWM can be written as~\cite{han2023learning}
\begin{equation}
\begin{split}
    \label{eq:metric_gbwm}
    {g}_{X}^{\mathrm{GBW}}&(S_{1},S_{2})=\frac{1}{2}{\mathrm{tr}}\left(\mathcal{L}_{X,M}\left(S_{1}\right) S_{2} \right) \\&=\frac{1}{2}\mathrm{vec}(S_{1})^{\top}(X\otimes M+M\otimes X)^{-1}\mathrm{vec}(S_{2}), 
\end{split}
\end{equation}
where $M\in\mathrm{Sym}^{+}_n$, $\mathcal{L}_{X,M}(S)$ is the generalized Lyapunov operator, which is the solution to the matrix equation ${X\mathcal{L}_{X,M}(S)M}+M{\mathcal{L}_{X,M}(S)X}={S}, S\in\mathrm{Sym}_n$. Choosing $M$ is equivalent to selecting a suitable metric. When $M=I_n$, it reduces to the BWM. When $M=X$, it coincides locally with the AIM~\cite{han2023learning}.
Consequently, it connects the BWM and AIM (locally). The corresponding  Riemannian distance is
\begin{equation}
\begin{split}
\label{eq:dist_gbwm}
    d_{\mathrm{GBW}}^{2}(X_{1},X_{2})=&\mathrm{tr}(M^{-1} X_{1})+\mathrm{tr}(M^{-1} X_{2}) \\&- 2\mathrm{tr}(X_{1}^{\frac{1}{2}}M^{-1}X_{2}M^{-1}X_{1}^{\frac{1}{2}})^{\frac{1}{2}}.
\end{split}
\end{equation}

\section{Proposed Method}
\label{sec:proposed_method}
The  proposed Riemannian self-attention networks (RieAtt) based on the $\theta$-GBWM can be derived from RieAtt based on the BWM, abbreviated as BWAtt. In this section, we first introduce BWAtt, and then elaborate on the generalization of BWAtt to a learnable version under $\theta$-GBWM.

\begin{figure*}[htbp]
\centering
\includegraphics[width=0.8\linewidth]{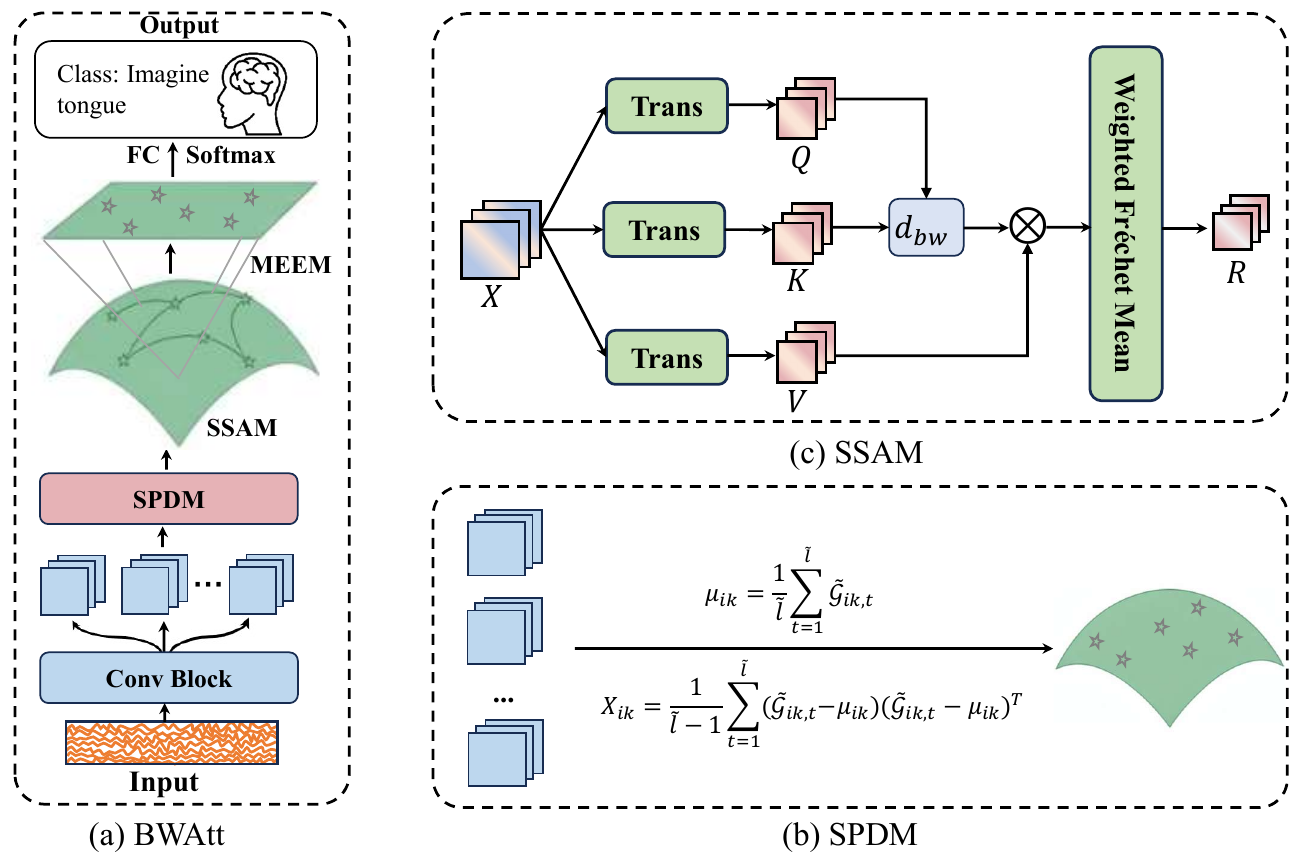} 
\caption{An overview of the proposed BWAtt. (a) illustrates the overall architecture of the BWAtt network; (b) shows the SPD modeling module; and (c) presents the SPD self-attention mechanism based on Bures-Wasserstein geometry.}
\label{fig:BW_att_network}
\end{figure*}

\subsection{RieAtt Based on the BWM}
\label{RASTT_BW}
As shown in \cref{fig:BW_att_network}, the network architecture of BWAtt is mainly constituted by the Spatiotemporal Feature Extraction Module (SFEM), the SPD Modeling Module (SPDM), the SPD Self-Attention Module (SSAM), and the Manifold to Euclidean Embedding Mapping (MEEM) layer. In the following, we give a detailed introduction to each of them.

\subsubsection{Spatiotemporal Feature Extraction Module (SFEM)}
\label{FE}

In the design of this module, we follow~\cite{sccnet} to make the SFEM contain two convolutional layers.
The first convolution performs spatial filtering on the multi-channel EEG signals with $N$ convolutional kernels. Specifically, given an input EEG signal $Y \in \mathbb{R}^{1 \times C \times T}$, where $C, T$ denote the number of channels and time samples, respectively, the spatial convolution with kernel size $(C,1)$ is defined as 
\begin{equation}
    Y^{(1)}_{n,1,t} = \sum_{i=0}^{C-1}Y_{1,1+i,t}P^{(1)}_{n,1,i,1} + b_{n}^{(1)},
\end{equation}
where  $P^{(1)} \in \mathbb{R}^{N \times 1 \times C \times 1}, b^{(1)} \in \mathbb{R}^{N}, n \in \{1, \cdots, N\}$ and $ t \in \{1, \cdots, T\}$.
The second convolution is applied to extract spatiotemporal features using $\tilde{N}$ convolutional kernels, with kernel size $(1,\tilde{T})$ and zero-padding of length $\tilde{T}/2$. After this operation, the data  after feature extraction is denoted as $Y^{(2)} \in \mathbb{R}^{\tilde{N} \times 1 \times (T+1)}$. \cref{tab:net_architecture_gbw} summarizes the changes of data dimensions after applying SFEM across different datasets.

\subsubsection{SPD Modeling Module (SPDM)}
For convenience, $\mathcal{G} \in \mathbb{R}^{n\times l}$ denotes the EEG spatiotemporal feature generated by SFEM.
To capture complementary statistical information embodied in different channel features, we first divide each $\mathcal{G}$ into $m$ non-overlapping sections along the channel dimension, denoted as $\{\tilde{\mathcal{G}}_{1},\tilde{\mathcal{G}}_{2},\cdots,\tilde{\mathcal{G}}_{m}\}\in\mathbb{R}^{n\times\tilde{l}}$, and then calculate the covariance matrix of each $\tilde{\mathcal{G}}_{i}, \forall i \in \{1,2,...,m\}$. The spatiotemporal information of $\mathcal{G}$ is encoded as a sequence of SPD matrices, expressed as $\{\Sigma_{1},\Sigma_{2},\cdots, \Sigma_{m}\} \in \mathrm{Sym}_{n}^{+}$. For each $\Sigma_{i} (i=1 \to m)$, we do trace-normalization, followed by the addition of a small perturbation $\epsilon$ on its diagonal elements. Now, the resulting SPD representation is signified as $\{X_{1}, X_{2}, \cdots, X_{m}\}$.

\subsubsection{SPD Self-Attention Module (SSAM)}
To capture the geometric relationship within $X_i$, we extend the Euclidean self-attention mechanism~\cite{vit} to the SPD manifold based on the BW geometry. The core operations of the proposed SSAM include:
SPD transformation layer, SPD attention layer, and SPD aggregation layer.

\noindent \textbf{SPD transformation layer.} Since the generated query, key, and value lie on the SPD manifold, conventional Euclidean transformations are not applicable. To this end, we adopt the bilinear transformation~\cite{spdnet} for structure-preserving mapping of SPD matrices. For a  sequence of SPD matrices $X_i \in \mathrm{Sym}_{n}^{+} (i = 1 \to m)$, we have: 
\begin{align}
    Q_{i} = f(X_i, W_q) = W_{q} X_{i} W_{q}^{\top},\\
    K_{i} = f(X_i, W_k) =W_{k} X_{i} W_{k}^{\top}, \\
    V_{i} =  f(X_i, W_v) =W_{v} X_{i} W_{v}^{\top},
\end{align}
where $W_q, W_k, W_v \in \mathbb{R}^{k\times n} (k<n)$ are learnable parameters of the row full-rank matrices lie in the Stiefel manifold. This transformation maps the SPD data from $\mathrm{Sym}^{+}_n$ to $\mathrm{Sym}^{+}_k$, while preserving the underlying geometric structure.

\noindent \textbf{SPD attention layer.} In Euclidean space, the inner product is commonly used to measure the similarity between query and key vectors. In our method, the query, key, and value matrices lie on the SPD manifold. Inspired by~\cite{pan2022matt}, we define similarity based on the geodesic distance induced by the BWM. Specifically, for $Q_i, K_j \in \mathrm{Sym}^{+}_k$, where $i,j \in \{1,\cdots, m\}$, each element of the attention matrix $\mathcal{A} = [a_{ij}]_{m \times m} $ is given by:
\begin{equation}
    a_{ij} = \frac{1}{1 + \log(1 + d_{\mathrm{BW}}(Q_i, K_j))}.
\end{equation}
To ensure that each row of the attention matrix $\mathcal{A}$ satisfies the convexity constraint, we apply the Softmax function along the row dimension to normalize the entries as follows:
\begin{equation}
    \tilde{a}_{ij} = \mathrm{Softmax}(\mathcal{A})=\frac{\exp(a_{ij})}{\sum_{j=1}^{m}\exp(a_{ij})}.
\end{equation}
Now, we have the final attention probability matrix $\tilde{\mathcal{A}}=[\tilde{a}_{ij}]_{m \times m}$. 

\noindent \textbf{SPD aggregation layer.} Let $X_1, \cdots, X_m$ denote m points on the SPD manifold. While it is plausible to treat the arithmetic mean $\frac{1}{m}\sum _{i\le {m}} {\boldsymbol{\mathrm{X}}_{i}}$ as the average of the SPD matrices $\{\boldsymbol{\mathrm{X}}_{i}\}_{i\le {m}}$, it is more appropriate to adopt Riemannian barycenter, also known as the Fréchet mean (FM)~\cite{yang2010riemannian}, from the perspective of respecting the Riemannian geometry. The theoretical and practical importance of the Riemannian barycenter in Riemannian data analysis has been well recognized, as evidenced in~\cite{PenX}, providing a valid rationale for its usage in this work. The FM of $\{\boldsymbol{\mathrm{X}}_{i}\}_{i\le {m}}$ is defined as:  
\begin{equation}
\label{mean_normal}
\begin{split}
    \mathrm{FM} (\{X_{i} \}) = \underset{G \in \mathrm{Sym}^{+}_k}{\mathrm{arg\,min}} \sum_{i=1}^{m}d_{\mathrm{BW}}^2\left(X_{i}, G \right).
\end{split}
\end{equation}
Eq. (\ref{mean_normal}) implies that the Riemannian barycenter is a point on the manifold that reduces to a minimum under the BWM. Similar to the weighted mean in Euclidean space, the above definition can also be extended to a weighted form. Let ${\Omega}=[{\omega}_1,\dots,{\omega}_m]$ be a weight vector, satisfying ${\omega }_{i}\ge {0}$ and $\sum_i {\omega}_{i}= 1$, the WFM can be expressed as:
\begin{equation}
\label{wfm}
        \mathrm{WFM} ( \{\omega_i\}, \{X_{i} \}) = \underset{G \in \mathrm{Sym}^{+}_k}{\mathrm{arg\,min}} \sum_{i=1}^{m}w_i d_{\mathrm{BW}}^2\left(X_{i}, G \right).
\end{equation}
Actually, \cref{wfm} admits a unique minimizer, which further corresponds to a solution of a nonlinear matrix equation~\cite{bhatia2019bures}:
\begin{equation}
\label{mean_equal}
    G=\sum_{i=1}^{m}\omega_{i}(G^{\frac{1}{2}}X_{i}G^{\frac{1}{2}})^{\frac{1}{2}}.
\end{equation}
When $m =2$, \cref{mean_equal} exists a closed-formed solution~\cite{bhatia2019bures}, which is given by:
\begin{equation}
\begin{split}
    \mathrm{WFM} ( \{\omega\}, &\{X_{i} \}) = (1-\omega)^{2}X_{1}+{\omega}^{2}X_{2}\\&+\omega(1-\omega)[(X_{1}X_{2})^{\frac{1}{2}}+(X_{2}X_{1})^{\frac{1}{2}}].
\end{split}
\end{equation}
When $m > 2$, we can not obtain the solution to \cref{mean_equal} in an explicit form, but it can be computed by fixed-point iteration~\cite{bhatia2019bures}:
\begin{equation}
    H(G)=G^{-\frac{1}{2}}\left(\sum_{i=1}^{m}\omega_{i}(G^{\frac{1}{2}}X_{i}G^{\frac{1}{2}})^{\frac{1}{2}}\right)^{2}G^{-\frac{1}{2}}.
\end{equation}
Its update rule is $G_{t+1}=H(G_{t})$. Although $G_{t}$ can be initialized as any arbitrary SPD matrix, it is reasonable to endow it with the arithmetic mean.
With the computed attention matrix $\tilde{\mathcal{A}}$ and a set of value matrices $V_{i} \in \mathrm{Sym}^{+}_k (i = 1 \to m)$, the $i$-th aggregated output $R_i \in \mathrm{Sym}^{+}_k$ is formulated as:
\begin{equation}
    R_i = \mathrm{WFM}(\{\tilde{\mathcal{A}}_{ij} \}_{j=1}^m, \{V_j \}_{j=1}^m).
\end{equation}
With these basic components in place, we summarize the forward procedure of the attention mechanism on the SPD manifold under the BWM into \cref{alg:coratt}.

\begin{algorithm}[htbp] \SetKwInOut{Input}{Input}\SetKwInOut{Output}{Output}\SetKwInOut{Parameters}{Parameters}
\caption{SPD Self-Attention Module under BWM}
\label{alg:coratt}
\Input{A sequence of SPD data $\{X_{1 \ldots m}\}$ \\ 
}
\Output{A sequence of SPD data $\{R_{1 \ldots m}\}$ \\
}
\BlankLine

\For{$i \leftarrow 1$ \KwTo $m$}{
    Queries: $Q_{i} = f(X_i, W_q) = W_{q} X_{i} W_{q}^{\top}$ \\
    Keys: $K_{i} = f(X_i, W_k) =W_{k} X_{i} W_{k}^{\top}$ \\
    Values: $V_{i} =  f(X_i, W_v) =W_{v} X_{i} W_{v}^{\top}$ \\
}

\For{$i \leftarrow 1$ \KwTo $m$}{
        \For{$j \leftarrow 1$ \KwTo $m$}{
            $\mathcal{A}_{ij} = \frac{1}{\left( 1 + \log(1 + d_{\mathrm{BW}}(Q_i, K_j)) \right)}$
        }
    Attention weight: $\tilde{\mathcal{A}}_{ij} = \operatorname{Softmax}(\mathcal{A}_{ij})$ \\
    Aggregation: $R_i = \operatorname{WFM}( \{\tilde{\mathcal{A}}_{ij} \}_{j=1}^m, \{ V_j \}_{j=1}^m)$ \\
}
\end{algorithm}

\subsubsection{Manifold to Euclidean Embedding Mapping (MEEM)} 
After passing through the SSAM, the ReEig layer is used to enforce nonlinearity while preserving the SPD structure. Unlike the ReLU function, which sets a threshold to the value of the input, the ReEig layer sets a threshold to the eigenvalues of the input~\cite{spdnet}. Formally, the ReEig function is defined as:
\begin{equation}
    \tilde{R}=f_{\mathrm{Re}}(R) = V \mathrm{max}(\epsilon I_k, \Sigma) V^{\top},
\end{equation}
where $R=V \Sigma V^{\top}$ is eigenvalue decomposition of $R$, $\epsilon>0$ is a small rectification threshold, $I_k$ is an $k \times k$ identity matrix. In this work, we set $\epsilon = 1e^{-5}$. Besides, $\mathrm{max}(\epsilon I_k, \Sigma)$ is a diagonal matrix with the following definition::
\begin{equation}
    \mathrm{max}(\epsilon I_k, \Sigma) =
    \begin{cases}
    \Sigma(i,i), &  \Sigma(i,i) > \epsilon,\\
    \epsilon, &  \Sigma_(i,i) \le \epsilon.
    \end{cases}
\end{equation}
To map the learned SPD representations back to the Euclidean space for enabling classification, we utilize the LogEig layer~\cite{spdnet} defined as follows:
\begin{equation}
    \hat{R}=f_{\mathrm{Log}}(\tilde{R})=\tilde{V} \mathrm{diag}(\log(\sigma_1),\cdots,\log(\sigma_k) \tilde{V}^{\top},
\end{equation}
where $\tilde{V}$ is the matrix of eigenvectors of  $\tilde{R}$. Then, we flatten the upper triangle of the symmetric matrix $\hat{R}$.
Finally, we set a fully connected layer and regular softmax operation for $\{\hat{R}_{1}, \hat{R}_{2}, \cdots, \hat{R}_{m} \}\in \mathrm{Sym}_k$. We use cross-entropy loss as the training objective.

\subsection{RieAtt Based on the GBWM}
\label{RASTT_GBW}
Recent research~\cite{thanwerdas2019affine,thanwerdas2022geometry,chen2024liebn} has extended AIM, LEM, LCM, and BWM into power-deformed families through the use of a matrix power diffeomorphism, that is $\phi_\theta(X)=X^\theta, \forall X \in \mathrm{Sym}^{+}_n$. Intuitively, this formulation allows interpolation between the original metric (when $\theta=1$) and a Log-Euclidean-like metric (as $\theta \to 0$)~\cite{thanwerdas2022geometry}. More recently,~\cite{wang2025learning} defines the power-deformed GBWM, denoted as $\theta$-GBWM, which can be expressed as:
{\small
\begin{equation} 
\label{eq:pow-metric}
    g^{\theta-\mathrm{GBW}}_{X} \left(S_{1},S_{2} \right) = \frac{1}{\theta^2} g_{X^\theta}^{\mathrm{GBW}} \left( \left(\phi_\theta\right)_{*,X} \left(S_{1}\right), \left(\phi_\theta\right)_{*,X} \left(S_{2}\right)\right),
\end{equation}}where $S_{1}, S_{2} \in {T}_{X}\mathrm{Sym}^{+}_n$, $\phi_\theta$ signifies the matrix power, and $(\phi_\theta)_{*,X}(\cdot)$ is the differential map of $\phi_\theta$ at $X$. As studied in~\cite{thanwerdas2022geometry}, the $\theta$-GBWM can be interpreted as a locally deformed AIM, and it approaches an LEM-like when $\theta \to 0$. These can be characterized by the following expressions:
{\small
\begin{equation}
    g^{\theta-\mathrm{GBW}}_{X} \left(S,S\right) \overset{\theta \to 0}{\longrightarrow}     
    \frac{1}{2} \left \langle \log_{*,X}\left(\mathcal{L}_{M}\left(S\right)\right),\log_{*,X}\left(S\right) \right \rangle,
\end{equation}}
\begin{equation}
    g^{\theta-\mathrm{GBW}}_{X} \left(S_{1},S_{2}\right)=\frac{1}{4}g^{\theta-\mathrm{AIM}}_{X}\left(S_{1},S_{2}\right).
\end{equation}
Therefore, the $\theta$-GBWM offers greater flexibility and may achieve better results than BWM when integrated with RieAtt. This is further demonstrated in the following experiments. 

As shown in~\cite{han2021generalized}, for any  $X \in \mathrm{Sym}^{+}_n$ and $M \in \mathrm{Sym}^{+}_n$, there exists a diffeomorphism: $\varphi: (\mathrm{Sym}^{+}_n,{g}^{BW}) \to (\mathrm{Sym}^{+}_n,{g}^{\theta-GBW})$, defined by $\varphi(X)=(M^{\frac{1}{2}}XM^{\frac{1}{2}})^\theta$. Therefore, the GBWM-based SSAM can be derived by applying the diffeomorphism to the one based on the BWM. The SPD transformation layer can be achieved in two steps: 1) The diffeomorphism to map SPD data from $(\mathrm{Sym}^{+}_n,{g}^{\mathrm{BW}})$ to $(\mathrm{Sym}^{+}_n,{g}^{\mathrm{\theta-GBW}})$. 2) The same bilinear transformation mapping as in BWM-based SSAM. Similarly, the WFM in the SPD aggregation layer can be defined by imposing the Riemannian isometry on the one based on the BWM, as shown below:
{\small
\begin{equation}
\label{iso_wfm}
    \widetilde{WFM} (\{\tilde{\mathcal{A}}_{ij} \}_{j=1}^m, \{V_j \}_{j=1}^m)=\varphi(\mathrm{WFM}( \{\tilde{\mathcal{A}}_{ij} \}_{j=1}^m, \{V_j \}_{j=1}^m)).
\end{equation}}

At this moment, we summarize the forward procedure of SPD Self-Attention Module under the $\theta$-GBWM  into \cref{alg:gBWAtt}.

\begin{algorithm}[htbp] \SetKwInOut{Input}{Input}\SetKwInOut{Output}{Output}\SetKwInOut{Parameters}{Parameters}
\caption{SPD Self-Attention Module under power deformed GBWM}
\label{alg:gBWAtt}
\Input{A sequence of SPD data $\{X_{1 \ldots m}\}$, power parameter $\theta \in \mathbb{R}\backslash\{0\}$, a learnable parameter $M\in \mathrm{Sym}^{+}_n$\\ 
}
\Output{A sequence of SPD data $\{R_{1 \ldots m}\}$ \\
}
\BlankLine

\For{$i \leftarrow 1$ \KwTo $m$}{
    Map from ${g}^{\mathrm{BW}}$ to ${g}^{\mathrm{{\theta}-GBW}}$: $\tilde{X}_i=(M^{\frac{1}{2}}XM^{\frac{1}{2}})^\theta$ \\
    Queries: $Q_{i} = f(\tilde{X}_i, W_q) = W_{q}\tilde{X}_{i} W_{q}^{\top}$ \\
    Keys: $K_{i} = f(\tilde{X}_i, W_k) =W_{k}\tilde{X}_{i} W_{k}^{\top}$ \\
    Values: $V_{i} =  f(\tilde{X}_i, W_v) =W_{v}\tilde{X}_{i} W_{v}^{\top}$ \\
}
\For{$i \leftarrow 1$ \KwTo $m$}{
        \For{$j \leftarrow 1$ \KwTo $m$}{
            $\mathcal{A}_{ij} = \frac{1}{\left( 1 + \log(1 + d_{\theta-\mathrm{GBW}}(Q_i, K_j)) \right)}$
        }
    Attention weight: $\tilde{\mathcal{A}}_{ij} = \operatorname{Softmax}(\mathcal{A}_{ij})$ \\
    Aggregation: $R_i = \widetilde{\operatorname{WFM}}( \{\tilde{\mathcal{A}}_{ij} \}_{j=1}^m, \{V_j \}_{j=1}^m)$ \\
}
\end{algorithm}
\section{Experiments}
\label{experiment}
In this section, we assess the performance of the proposed GBWAtt in three typical EEG tasks, which are the MI decoding on the BCIC-IV-2a dataset~\cite{brunner2008bci}, SSVEP decoding on the MAMEM-SSVEP-II dataset~\cite{mamem}, and Error-Related Negativity (ERN) decoding on the BCI-ERN dataset~\cite{bciern}. To ensure a comprehensive comparison, we include several baseline models based on conventional Euclidean deep learning, such as ShallowConvNet~\cite{matt18}, EEGNet~\cite{matt17}, SCCNet~\cite{sccnet}, MBEEGSE~\cite{matt50}, TCNet-Fusion~\cite{matt51},  and FBCNet~\cite{matt53}. Moreover, we incorporate several state-of-the-art geometric deep learning models, including SPDNet~\cite{spdnet}, SPDNetBN~\cite{spdnetbn}, MAtt~\cite{pan2022matt}, and GDLNet~\cite{gdlnet}, to provide a more convincing comparison. All experiments were conducted on an Intel i9-13900HX CPU with 16GB of RAM and an NVIDIA RTX4060 GPU.

\subsection{Datasets and Experimental Setting}
\label{dataset}
\textbf{MI.} The BCIC-IV-2a dataset is a representative public EEG resource for time-asynchronous EEG decoding, containing signals from 9 subjects performing a four-class MI task. Each subject completed two sessions conducted on different days, with each trial involving four seconds of imagined movement (right hand, left hand, feet, or tongue). Each session includes 288 trials, with 72 trials per class. Following the protocol in~\cite{pan2022matt}, the first session of BCIC-IV-2a is used for training, reserving one-eighth of it for validation. Besides, the performance indicator is based on classification accuracy.

\textbf{SSVEP.} MAMEM-SSVEP-II dataset includes time-synchronized EEG data recordings from 11 subjects, each contributing five sessions. In each session,  participants were asked to concentrate on any one of five distinct visual
stimuli flickering at different frequencies: 6.66, 7.50, 8.57, 10.00, or 12.00 Hz. Each subject completed five trials, one for each frequency, yielding 100 trials per session. Each trial lasted between 1 to 5 seconds after the prompt, divided into four one-second segments. To ensure an equitable comparison, we adhere to the data preparation and performance evaluation procedures described in~\cite{pan2022matt,gdlnet}. Specifically, the first four sessions of the subject are used as the training set, with session 4 allocated for validation.

\textbf{ERN.} The BCI-ERN dataset originates from a BCI Challenge on Kaggle. It contains EEG recordings from 26 subjects who participated in a P300-based spelling task, designed to measure ERN. Since ERN responses are triggered only by incorrect outputs, the resulting classification task is inherently binary and imbalanced, with correct inputs significantly outnumbering erroneous ones. The objective is to detect signal perturbations associated with errors, thereby evaluating and enhancing model robustness. Following the protocol in~\cite{pan2022matt,gdlnet}, we adopt the same dataset pre-processing as in the MAMEM-SSVEP-II dataset and report model performance using the Area Under the Curve (AUC) metric.

\begin{table}[t]
\caption{GBWAtt architectures across three datasets, where SpatConv and SpatTempConv denote spatial and spatiotemporal convolution layers. The Attention block represents the SPD Self-Attention Module.}
\centering
\footnotesize
\begin{tabular}{lcccl}
\toprule
\textbf{Block} & \textbf{MI}  & \textbf{MAMEM}  & \textbf{ERN}  & \\
\midrule
Input data              & $1 \times 22 \times 438$  & $1 \times 8 \times 125$       & $1 \times 56 \times 160$ \\
SpatConv                & $22 \times 1 \times 438$  & $125 \times 1 \times 125$     & $14 \times 1 \times 160$ \\
SpatTempConv            & $20 \times 1 \times 439$  & $15 \times 1 \times 126$      & $42 \times 1 \times 161$ \\
Split                   & $3 \times 20 \times 20$  & $7 \times 15 \times 15$       & $3 \times 16\times 16$ \\
Attention           & $3 \times 18 \times 18 $  & $7 \times 12 \times 12$       & $3 \times 8 \times 8$ \\
LogEig        & $3 \times 18 \times 18 $  & $7 \times 12 \times 12$       & $3 \times 8 \times 8$ \\
Vectorization           & $513$                     & $546$                         & $108$ \\
FC + Softmax            & $4$                       & $5$                           & $2$ \\
\bottomrule
\end{tabular}
\label{tab:net_architecture_gbw}
\end{table}

\textbf{Implementation Details.} All SPD parameters are optimized using the Geoopt~\cite{kochurov2020geoopt}. For the BCIC-IV-2a dataset, the number of subparts, the size of the transformation matrix in GBWAtt, the learning rate, and the batch size are respectively set to $3$, $20\times 18$, $2.5\mathrm{e}^{-3}$, and $128$, respectively. For the MAMEM-SSVEP-II dataset, these parameters are configured as $7$, $15\times 12$, $1\mathrm{e}^{-3}$, and $64$, respectively. In the case of the BCI-ERN dataset, they are set to $3$, $16\times 8$, $5\mathrm{e}^{-4}$, and $32$. For these three datasets, the number of training epochs is set to $350$, $180$, and $130$, respectively. For clarity and reproducibility, the detailed network configurations of GBWAtt across all datasets are summarized in \cref{tab:net_architecture_gbw}. Since GBWAtt involves several manifold-valued matrix operations, we further provide its computational complexity and practical training efficiency analysis in~\cref{appendix:complexity}.

\begin{table}[t] 
\caption{Average performance ($\pm$ standard deviation) over 10 runs, comparing GBWAtt with SOTA methods on three EEG datasets.}
\centering
\resizebox{\linewidth}{!}
{
\begin{tabular}{l c c c}
\toprule
\textbf{Models} &  \textbf{MI} & \textbf{SSVEP} & \textbf{ERN} \\
\midrule
EEGNet ~\cite{matt17}       & 61.84 $\pm$ 6.39 & 53.72 $\pm$ 7.23  & 74.28 $\pm$ 2.47 \\
ShallowCNet ~\cite{matt18}  & 57.43 $\pm$ 6.25 & 56.93 $\pm$ 6.97  & 71.86 $\pm$ 2.64 \\
SCCNet ~\cite{sccnet}       & 71.95 $\pm$ 5.05 & 62.11 $\pm$ 7.70  & 70.93 $\pm$ 2.31 \\
FBCNet  ~\cite{matt53}      & 56.52 $\pm$ 3.07 & 53.09 $\pm$ 5.67  & 60.47 $\pm$ 3.06 \\
TCNet-Fusion ~\cite{matt51} & 71.45 $\pm$ 4.45 & 45.00 $\pm$ 6.45  & 70.46 $\pm$ 2.94 \\
MBEEGSE  ~\cite{matt50}     & 64.58 $\pm$ 6.07 & 56.45 $\pm$ 7.27  & 75.46 $\pm$ 2.34 \\
SPDNet   ~\cite{spdnet}     & 72.93 $\pm$ 4.33 & 62.30 $\pm$ 3.12  & 72.05 $\pm$ 4.43 \\
SPDNetBN ~\cite{spdnetbn}   & 73.02 $\pm$ 3.67 & 62.76 $\pm$ 3.01  & 72.34 $\pm$ 3.46 \\
MAtt   ~\cite{pan2022matt}  & 74.71 $\pm$ 5.01 & 65.19 $\pm$ 3.14  & 75.68 $\pm$ 2.23 \\
GDLNet ~\cite{gdlnet}       & 69.32 $\pm$ 2.89 & 65.52 $\pm$ 2.86  & 78.23 $\pm$ 2.52 \\
\midrule
GBWAtt ($\theta=1$) & 74.54 $\pm$ 2.04 & 67.12 $\pm$ 2.94 & 80.42 $\pm$ 2.12 \\
GBWAtt ($\theta=1.5$) & \textbf{74.95 $\pm$ 1.83} & \textbf{67.27 $\pm$ 2.62} & \textbf{{80.90 $\pm$ 1.97}} \\
\bottomrule
\end{tabular}
}
\label{tab:result_main}
\end{table}

\subsection{Result}
\label{result}
\cref{tab:result_main} lists the experimental results of GBWAtt and the selected comparison methods on the three used EEG datasets under 10-fold validation. Here, we report the performance of GBWAtt using both the standard metric ($\theta=1$) and the deformed metric ($\theta=1.5$). 
It can be seen that Riemannian Neural Networks (RieNets) consistently outperform most Euclidean deep learning-based EEG models, highlighting the effectiveness of Riemannian geometry in encoding the nonlinear structure of sequential signals. Notably, on the BCIC-IV-2a dataset, modeling data on the SPD manifold (as implemented in SPDNet, SPDNetBN and MAtt) yields significant performance gains compared to modeling on the Grassmann manifold (as in GDLNet). Furthermore, the proposed GBWAtt achieves the best performance across these three tasks, further validating its effectiveness in geometric data analysis. In particular, GBWAtt surpasses MAtt with performance improvements of approximately 0.24\% on the MI dataset, 2.08\% on SSVEP, and 5.22\% on ERN, demonstrating the advantage of the proposed self-attention mechanism under the GBWM. Additionally, it can also be noted that the GBWAtt generally benefits from power deformation, as the deformed metric ($\theta=1.5$) outperforms the standard metric ($\theta=1$) on all datasets, emphasizing the importance of metric parameterization.

\begin{table}[htbp] 
\caption{Ablation of major modules in the proposed model.}
\centering
\resizebox{\linewidth}{!}{
\begin{tabular}{lccc}
\toprule
\textbf{Models} & \textbf{MI} & \textbf{SSVEP} & \textbf{ERN} \\
\midrule
SFEM & $26.22 \pm 0.84$ & $20.26 \pm 1.23$ & $73.22 \pm 2.63$ \\
GBW-SSAM & $55.96 \pm 3.58$ & $30.18 \pm 2.43$ & $58.10 \pm 3.24$ \\
SFEM + ESAM & $49.42 \pm 2.15$ & $22.97 \pm 2.33$ & $63.95 \pm 2.44$ \\
\midrule
SFEM + GBW-SSAM & $\textbf{74.95} \pm \textbf{1.83}$ & $\textbf{67.27} \pm \textbf{2.62}$ & $\textbf{80.90} \pm \textbf{1.97}$ \\
\bottomrule
\end{tabular}}
\label{tab:ablation_component}
\end{table}

\subsection{Ablation}
\label{ablation}

\textbf{Ablation of major modules.}
As shown in \cref{tab:ablation_component}, the exclusion of any GBWAtt component results in a drop in model accuracy, underscoring the importance of each part. Furthermore, integrating the Euclidean Self-Attention Module (ESAM) after the Spatiotemporal Feature Extraction Module (SFEM)  also degrades performance, further validating the superiority of the SPD Self-Attention Module based on the $\theta$-GBWM (GBW-SSAM).

\begin{table}[htbp]
\caption{Component-wise ablations on MAMEM and ERN.}
\centering
\resizebox{\linewidth}{!}{
\begin{tabular}{llcc}
\toprule
\textbf{Component} & \textbf{Design} & \textbf{MAMEM} & \textbf{ERN} \\
\midrule
\multirow{4}{*}{Metric}
& Euclidean & $56.55 \pm 2.91$ & $63.95 \pm 3.24$ \\
& AIM & $64.39 \pm 2.74$ & $76.42 \pm 2.78$ \\
& BWM & $65.85 \pm 2.54$ & $79.82 \pm 2.24$ \\
& GBWM & $\textbf{67.27} \pm \textbf{2.62}$ & $\textbf{80.90} \pm \textbf{1.97}$ \\
\midrule
\multirow{4}{*}{Attention score}
& Inner product & $63.72 \pm 3.15$ & $75.37 \pm 2.62$ \\
& Gaussian kernel & $64.23 \pm 2.75$ & $76.62 \pm 2.48$ \\
& Negative squared geodesic distance & $63.54 \pm 2.49$ & $75.42 \pm 2.64$ \\
& Inverse-log distance & $\textbf{67.27} \pm \textbf{2.62}$ & $\textbf{80.90} \pm \textbf{1.97}$ \\
\midrule
\multirow{3}{*}{Aggregation}
& Euclidean averaging & $56.55 \pm 3.07$ & $70.62 \pm 2.48$ \\
& Tangent-space averaging & $61.21 \pm 2.91$ & $76.42 \pm 2.64$ \\
& Fr\'echet mean aggregation & $\textbf{67.27} \pm \textbf{2.62}$ & $\textbf{80.90} \pm \textbf{1.97}$ \\
\midrule
\multirow{2}{*}{MEEM}
& GBWAtt w/o MEEM & $59.46 \pm 2.35$ & $68.52 \pm 4.27$ \\
& GBWAtt & $\textbf{67.27} \pm \textbf{2.62}$ & $\textbf{80.90} \pm \textbf{1.97}$ \\
\bottomrule
\end{tabular}}
\label{tab:ablation_key_components}
\end{table}

\textbf{Ablation of GBWAtt designs.}
To further isolate the contribution of each design choice in GBWAtt, we conduct fine-grained ablation studies from four aspects: metric choice, attention score, feature aggregation, and the manifold-to-Euclidean embedding mapping (MEEM). For metric design, we compare Euclidean, AIM, BWM, and the proposed GBWM. For attention scoring, we compare four similarity functions, including inner product, Gaussian kernel, negative squared geodesic distance, and the proposed inverse-log distance. For feature aggregation, we compare Euclidean averaging, tangent-space averaging, and Fr\'echet mean aggregation.

For the inner-product-based similarity, since the standard Euclidean dot product is not directly available on the SPD manifold, we follow prior Riemannian formulations~\cite{nguyen2023building} and first map SPD matrices to the tangent space at the identity. The induced tangent-space inner product is defined as
\begin{equation}
    \langle P,Q\rangle
    =
    \left\langle \operatorname{Log}_{I_d}(P),
    \operatorname{Log}_{I_d}(Q)
    \right\rangle_{I_d},
    \quad P,Q\in\mathrm{Sym}^{+}_n.
\end{equation}

As shown in \cref{tab:ablation_key_components}, GBWM achieves the best performance among different metric choices, outperforming Euclidean, AIM, and BWM on both datasets. This confirms the effectiveness of the learnable power deformation built upon the Bures--Wasserstein geometry. In terms of attention scoring, the inverse-log distance consistently outperforms the inner product, Gaussian kernel, and negative squared geodesic distance. In particular, geodesic-distance-based similarities generally perform better than the tangent-space inner-product similarity. This may be because the tangent-space inner product provides a local linear approximation around the reference point, whereas geodesic distances better capture the intrinsic relationships between SPD representations. Moreover, Fr\'echet mean aggregation performs better than Euclidean and tangent-space aggregation, confirming the importance of preserving the intrinsic geometry of SPD features during aggregation. Finally, removing MEEM leads to a clear performance drop, which verifies its role in preserving geometrical information before downstream Euclidean classification.

\begin{table}[htbp]
\centering
\caption{Accuracy (\%) under different $\theta$.}
\label{tab:theta_sensitivity}
\resizebox{\linewidth}{!}{
\begin{tabular}{lccccccc}
\toprule
Dataset & 0.25 & 0.5 & 0.75 & 1 & 1.25 & 1.5 & 2 \\
\midrule
MAMEM & $65.57 \pm 3.12$ & $66.27 \pm 2.75$ & $66.79 \pm 3.42$ & $67.12 \pm 2.62$ & $67.04 \pm 3.12$ & $67.27 \pm 2.62$ & $67.03 \pm 2.21$ \\
ERN & $78.46 \pm 2.12$ & $79.25 \pm 2.16$ & $80.72 \pm 2.24$ & $80.08 \pm 1.79$ & $80.14 \pm 2.08$ & $80.90 \pm 1.97$ & $79.74 \pm 2.32$ \\
\bottomrule
\end{tabular}}
\end{table}

\textbf{Effect of the deformation parameter $\theta$.}
We further investigate the effect of the deformation parameter $\theta$ in the proposed GBWM by varying $\theta \in \{0.25, 0.5, 0.75, 1, 1.25, 1.5, 2\}$. As shown in Table~\ref{tab:theta_sensitivity}, very small values of $\theta$ are generally suboptimal, whereas moderate values lead to better performance. In particular, the accuracy on MAMEM improves from $65.57\%$ at $\theta=0.25$ to $67.27\%$ at $\theta=1.5$, and ERN also achieves its best result at $\theta=1.5$. These results demonstrate that the power deformation is effective and a moderate deformation strength can better adapt the underlying Bures-Wasserstein geometry to EEG representation learning.

\textbf{Evaluation on large-scale clinical EEG datasets.}
To further evaluate the generalization ability and practical applicability of the proposed GBWAtt, we conduct additional experiments on two large-scale clinical EEG datasets, TUAB and TUEV. Following the SPD modeling strategy in ~\cite{Rhop}, we integrate GBWAtt into two representative EEG foundation models, BIOT~\cite{Biot} and LaBraM~\cite{LaBraM}, and compare it with both the original pretrained backbone and the MAtt-enhanced variant. As reported in \cref{tab:clinical_eeg_results}, GBWAtt consistently improves the corresponding backbone and outperforms MAtt, demonstrating that the proposed GBWAtt is not limited to conventional EEG benchmarks and can be effectively integrated into modern large-scale EEG architectures. 

\begin{table}[htbp]
\caption{Balanced accuracy on TUAB and TUEV.}
\centering
\resizebox{0.85\linewidth}{!}{
\begin{tabular}{lcc}
\toprule
\textbf{Model} & \textbf{TUEV} & \textbf{TUAB} \\
\midrule
BIOT-Base~\cite{Biot} & $0.5281 \pm 0.0225$ & $0.7959 \pm 0.0057$ \\
BIOT-MAtt & $0.5304 \pm 0.0218$ & $0.7972 \pm 0.0083$ \\
BIOT-GBWAtt & $\textbf{0.5362} \pm \textbf{0.0154}$ & $\textbf{0.8016} \pm \textbf{0.0034}$ \\
\midrule
LaBraM-Base~\cite{LaBraM} & $0.6409 \pm 0.0065$ & $0.8140 \pm 0.0019$ \\
LaBraM-MAtt & $0.6356 \pm 0.0072$ & $0.8162 \pm 0.0016$ \\
LaBraM-GBWAtt & $\textbf{0.6504} \pm \textbf{0.0045}$ & $\textbf{0.8190} \pm \textbf{0.0014}$ \\
\bottomrule
\end{tabular}}
\label{tab:clinical_eeg_results}
\end{table}

\begin{figure}[htbp]
\centering
\includegraphics[width=0.7\linewidth]{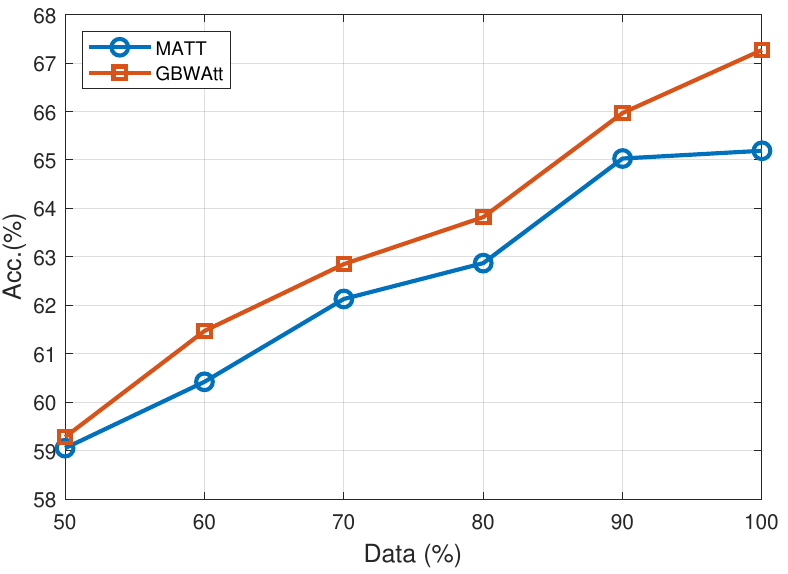} 
\caption{Comparison of the limited data on the MAMEM.}
\label{fig:limited}
\end{figure}

\textbf{Comparison on the limited data.} To evaluate the performance of our method under limited data conditions, we conduct experiments on the MAMEM dataset. Specifically, we progressively reduce the size of the training set while keeping the validation and test sets unchanged, considering training set proportions of {50\%, 60\%, 70\%, 80\%, 90\%, and 100\%}. As shown in \cref{fig:limited}, our method consistently outperforms MAtt across all levels of training data reduction, demonstrating its superior robustness and effectiveness even when training data is limited.

\textbf{Comparison on model robustness.}
Here, we further evaluate the robustness of the proposed model by introducing Gaussian noise to the input data and comparing it with MAtt~\cite{pan2022matt} on the MAMEM dataset. The noise $\mathcal{N}(\mu,\sigma^2)$ is generated with mean $\mu=0$ and variance $\sigma^2 \in \{0.2,0.4,0.6,0.8,1.0\}$. As illustrated in \cref{tab:ablation_robustness}, our method consistently outperforms MAtt across all noise levels, further demonstrating its robustness. 

\begin{table}[htbp] 
\caption{Comparison under different Gaussian noise levels on the MAMEM dataset.}
\centering
\resizebox{\linewidth}{!}
{
\begin{tabular}{l c c c c c}
\toprule
Gaussian Noise & $\mathcal{N}$(0, 0.2) & $\mathcal{N}$(0, 0.4) & $\mathcal{N}$(0, 0.6) & $\mathcal{N}$(0, 0.8) & $\mathcal{N}$(0, 1.0) \\
\midrule
MAtt    &  63.52 $\pm$ 2.56 & 63.22 $\pm$ 2.74 & 62.25 $\pm$ 3.18 & 60.31 $\pm$ 3.01 & 59.85 $\pm$ 2.76 \\
GBWAtt &  \textbf{66.69 $\pm$ 2.18} & \textbf{64.61 $\pm$ 2.60} & \textbf{62.71 $\pm$ 2.31} & \textbf{60.51 $\pm$ 2.57} & \textbf{60.31 $\pm$ 2.13} \\
\bottomrule
\end{tabular}
}
\label{tab:ablation_robustness}
\end{table}

To better understand this robustness improvement, we analyze the proportion of ill-conditioned SPD matrices before attention, after MAtt, and after GBWAtt under different Gaussian noise levels. Following the common condition-number criterion, an SPD matrix is regarded as ill-conditioned if $\kappa=\lambda_{\max}/\lambda_{\min}>10^4$. As reported in \cref{tab:condition_number_noise}, the proportion of ill-conditioned matrices before attention increases sharply from $54.77\%$ under $\mathcal{N}(0,0.2)$ to $91.09\%$ under $\mathcal{N}(0,1.0)$, indicating that strong noise destroys the intrinsic covariance structure of the input data. After the attention module, GBWAtt consistently produces fewer ill-conditioned SPD features than MAtt across all noise levels. For example, under $\mathcal{N}(0,1.0)$, the proportion of ill-conditioned matrices is reduced to $7.46\%$ by GBWAtt, compared with $14.27\%$ by MAtt. These results suggest that GBWAtt can better alleviate numerical instability caused by noisy covariance representations. Meanwhile, when the input covariance structure is heavily corrupted by very strong noise, both methods operate on less informative SPD representations, which explains why their performance gap becomes smaller at high noise levels.

\begin{table}[htbp]
\caption{Statistics of ill-conditioned SPD features ($\kappa=\lambda_{\max}/\lambda_{\min}>10^4$) under different Gaussian noise levels on the MAMEM dataset.}
\centering
\resizebox{\linewidth}{!}{
\begin{tabular}{lccc}
\toprule
Gaussian Noise & Before Attention & After GBWAtt & After MAtt \\
\midrule
$\mathcal{N}(0,0.2)$ & $54.77\%$ & $0.00\%$ & $1.42\%$ \\
$\mathcal{N}(0,0.4)$ & $60.49\%$ & $0.40\%$ & $3.16\%$ \\
$\mathcal{N}(0,0.6)$ & $75.03\%$ & $1.50\%$ & $6.74\%$ \\
$\mathcal{N}(0,0.8)$ & $84.34\%$ & $5.42\%$ & $10.42\%$ \\
$\mathcal{N}(0,1.0)$ & $91.09\%$ & $7.46\%$ & $14.27\%$ \\
\bottomrule
\end{tabular}}
\label{tab:condition_number_noise}
\end{table}

\begin{figure*}[t]
\centering
\includegraphics[width=0.8\linewidth]{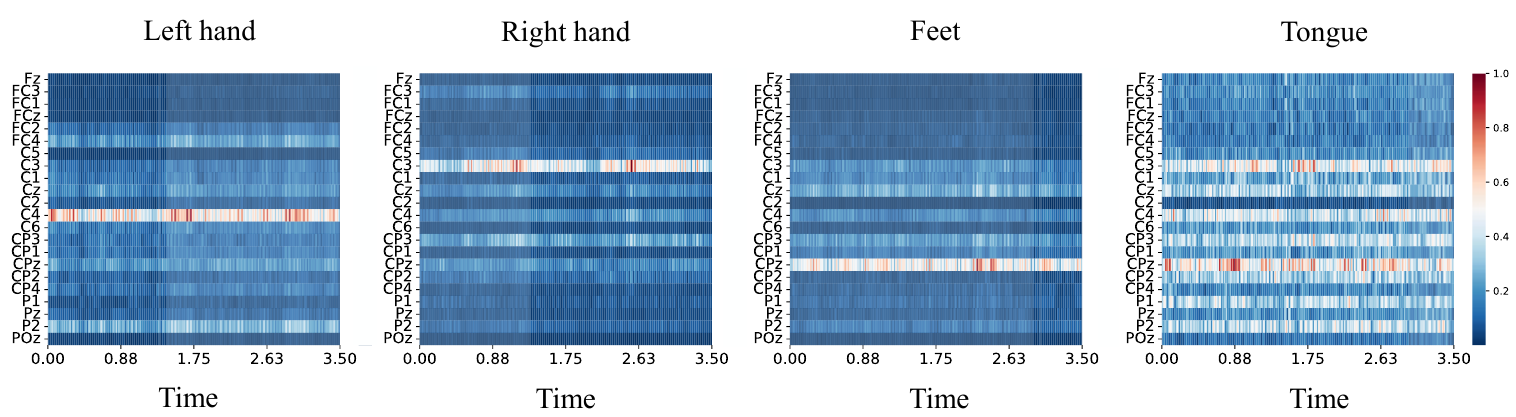} 
\caption{Heatmaps of GBWAtt for the S3 subject across four motor-imagery classes on the BCIC-IV-2a dataset. The x-axis and y-axis represent time and EEG channels, respectively.}
\label{fig:bci_heat}
\end{figure*} 

\begin{figure*}[t]
\centering
\includegraphics[width=0.8\linewidth]{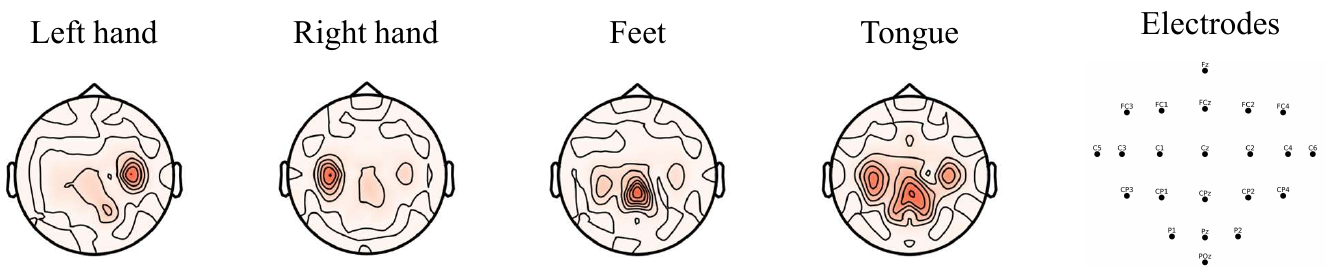} 
\caption{The spatial topo-maps and the diagram of electrode distribution of GBWAtt for the S3 subject across four motor-imagery classes on the BCIC-IV-2a dataset.}
\label{fig:bci_topo}
\end{figure*}

\subsection{EEG Model Interpretation}
\label{model interpretation}

In \cref{fig:bci_heat,fig:bci_topo}, we present the gradient responses of subject S3 during MI tasks, showcasing how GBWAtt interprets EEG data across both spatial channels and temporal progression. For left-hand and right-hand MI, contralateral channels C4 and C3 exhibit prominent activity: left-hand imagery maintains strong gradients throughout the trial, whereas right-hand imagery shows distinct peaks at approximately 1.0–1.4 seconds and 2.4–2.8 seconds. In the case of feet MI, significant activation appears at the midline channel CPz within the motor cortex, primarily between 2.0 and 2.6 seconds. For tongue MI, substantial activations are observed at C3 and CPz in the early phase (0.0–1.0 seconds). These heatmaps present absolute gradient magnitudes across EEG channels (y-axis) and time (x-axis) for the four MI classes (left hand, right hand, feet, tongue) in the BCIC-IV-2a dataset for S3. Red regions indicate areas of strong gradient response. The findings highlight that C4 is most informative for left-hand imagery, C3 for right-hand imagery, CPz for feet imagery, and both C3 and CPz for tongue imagery, demonstrating how distinct EEG channels play key roles in discriminating between MI classes. The detailed comparison is available at~\cref{app:add_interp}.

\section{Conclusion}
\label{Sec:Conclusion}
In this paper, we introduce a novel SPD self-attention mechanism based on the BWM for EEG decoding. Furthermore, we extend this approach to incorporate a $\theta$-GBWM. Extensive experiments and ablation studies across three EEG tasks demonstrate the effectiveness and superiority of our method compared to other Riemannian neural networks. The proposed GBWAtt framework is expected to contribute to Riemannian deep learning for EEG processing, offering enhanced efficiency and robustness.

\section*{Limitations and Ethical Considerations}
The proposed method incurs additional computational cost due to repeated manifold-valued matrix operations under Bures-Wasserstein geometry. Moreover, although it is more robust than the baselines, its relative advantage becomes smaller under very strong noise, where the covariance structure is heavily corrupted. From an ethical perspective, this work uses standard public and de-identified EEG datasets and is intended as a methodological study rather than a clinical decision system. Therefore, it does not introduce additional risks beyond those commonly associated with biomedical signal analysis and EEG-based pattern recognition.

\section*{GenAI Disclosure}
During the preparation of this manuscript, we used generative AI tools to assist with language polishing and LaTeX formatting. All technical content, mathematical formulations, experimental design, and experimental results were fully conceived and validated by the authors.

\begin{acks}
This work was supported in part by the National Natural Science Foundation of China (62306127, 62576153, 62332008), the Natural Science Foundation of Jiangsu Province (BK20231040), Postgraduate Research \& Practice Innovation Program of Jiangsu Province (22671065), the Fundamental Research Funds for the Central Universities (JUSRP124015), and the National Key R\&D Program of China (2023YFF1105102, 2023YFF1105105).
\end{acks}

\bibliographystyle{ACM-Reference-Format}
\bibliography{sample-base}

\clearpage
\appendix

\section{Related Work}
\label{app:related}
\subsection{Neural Networks for EEG}
Due to the superior feature extraction ability of deep learning, numerous end-to-end models have been proposed for EEG signal decoding~\cite{an2023dual}. To name a few, EEGInception~\cite{santamaria2020eeg} employs a ConvNet architecture based on temporal inception modules for MI classification. Similarly, ConvDCDNN~\cite{zhang2024convolutional} improves EEG classification with minimal human intervention. 
Inspired by the brain’s cognitive mechanisms, Li et al.~\cite{li2024efficient} proposed a graph learning framework based on GCN to improve the performance of EEG-based emotion recognition. Although these methods have made significant progress in EEG decoding, they often fail to consider the inherent non-Euclidean geometric properties of EEG data. To address this issue, Suh et al.~\cite{suh2021riemannian} employed SPDNet to extract statistical features from EEG signals and introduced a separate-to-learn strategy. Pan et al.~\cite{pan2022matt} introduced MAtt, an SPD attention network leveraging the LEM to represent EEG signals as SPD matrices, and Wang et al.~\cite{gdlnet} proposed a novel Grassmannian Attention Network, which models channel features as Grassmann subspaces and implements self-attention mechanisms within the manifold space.

\begin{figure*}[t]
\centering
\includegraphics[width=0.9\linewidth]{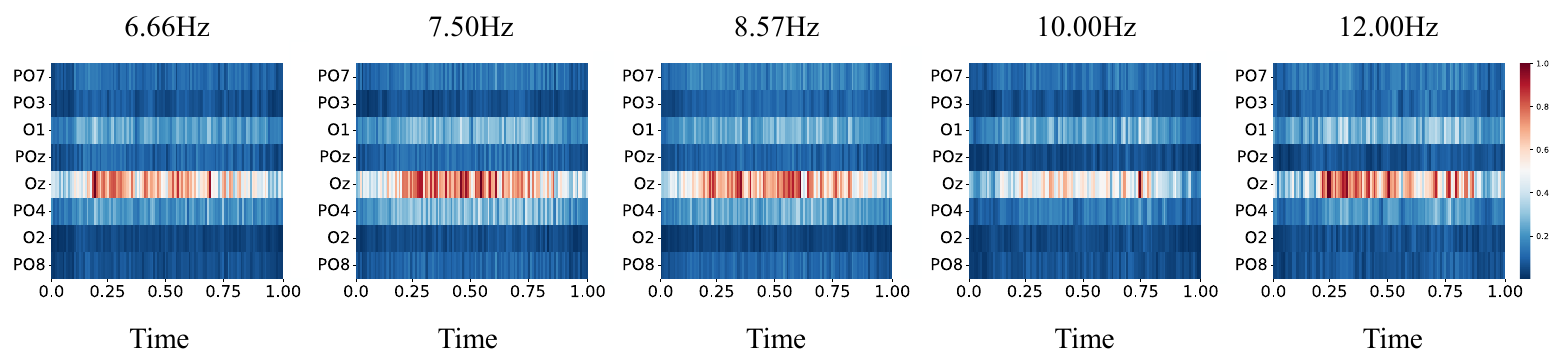} 
\caption{Heatmaps of GBWAtt for the S11 subject across five different frequencies on the  MAMEM-SSVEP-II dataset. The x-axis and y-axis represent time and EEG channels, respectively.}
\label{fig:mamem_heat}
\end{figure*}

\begin{figure*}[t]
\centering
\includegraphics[width=0.9\linewidth]{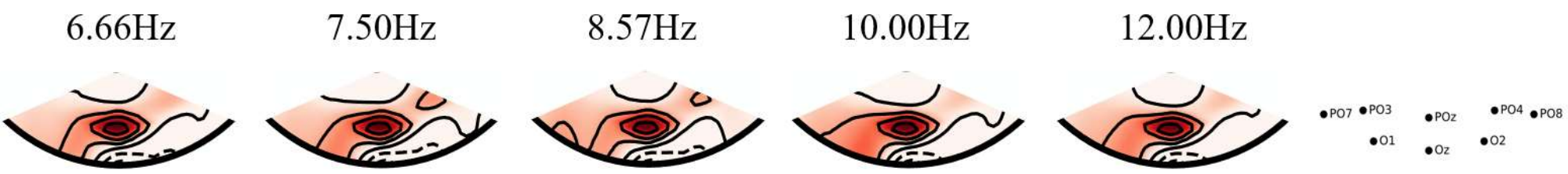} 
\caption{The spatial topo-maps and the diagram of electrode distribution  of GBWAtt for the S11 subject across five different frequencies on the SSVEP dataset.}
\label{fig:mamem_topo}
\end{figure*} 

\begin{figure*}[t]
\centering
\includegraphics[width=0.9\linewidth]{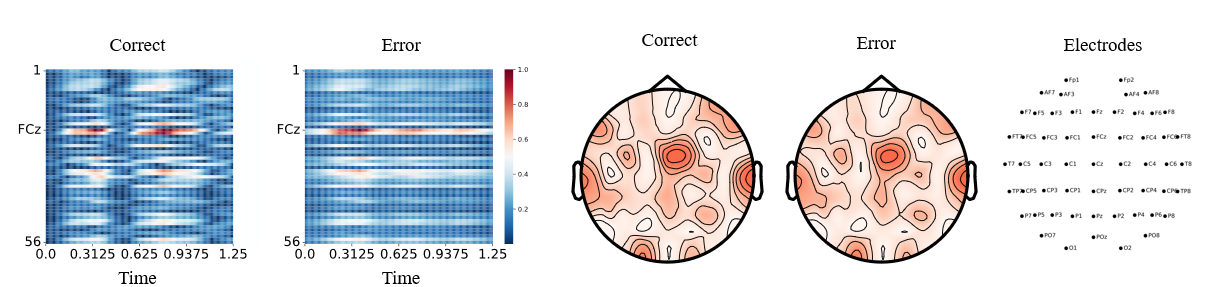} 
\caption{The heatmaps and the visualization result of GBWAtt for
two classes on the BCI-ERN datasets S7 model. The x-axis represents time,
and the y-axis represents EEG channels.}
\label{fig:ern}
\end{figure*} 

\subsection{Riemannian Networks for SPD Matrices}
Inspired by the superior feature extraction capability of ConvNets, Huang et al.~\cite{spdnet} proposed an end-to-end Riemannian neural network. Differently, this network is designed to process structured data, enabling nonlinear, end-to-end learning on SPD matrices.  On this basis, Nguyen et al.~\cite{hgrnet} designed two aggregation subnetworks in the context of Gaussian embedded Riemannian manifolds for effectively encoding and learning spatial and temporal correlations. Zhang et al.~\cite{dmtnet} recognized the importance of local structural information for learning discriminative representations and introduced a 2D convolutional layer operating on SPD features, with convolution kernel matrix constrained to be SPD. Considering that this method does not respect the geometric structure of the data, Chakraborty et al.~\cite{manifoldnet} further proposed to simulate Euclidean convolution by calculating the weighted Fréchet mean over manifold-valued data grid, where each grid element corresponds to an SPD matrix, achieving better preservation of manifold properties. However, this approach does not explicitly capture the statistics within individual grid elements. Chen et al.~\cite{msnet} adhered more rigorously to the principles of Riemannian geometry by leveraging the category theory to abstract local structures on the SPD manifold, ultimately developing a multi-scale network for capturing local geometric features.  Moreover, to enhance the applicability of Riemannian networks in visual tasks with limited training data, Wang et al.~\cite{symnet} developed a lightweight Riemannian network for SPD matrix nonlinear learning. This network employs a shallow optimization strategy for parameter learning, thereby improving computational efficiency while maintaining high classification accuracy.

\section{Computational Complexity and Training Efficiency}
\label{appendix:complexity}

All operations in GBWAtt are implemented using explicit formulations or standard iterative procedures. Specifically, the Lyapunov operator is computed via SVD-based matrix functions, the weighted Fréchet mean is solved by a standard fixed-point iteration, and the metric deformation is realized through matrix power operations. Therefore, the computational complexity of GBWAtt is $O(aNd^3)$, where $N$, $d$, and $a$ denote the batch size, the matrix dimension, and the number of SVD-based matrix functions, respectively. Compared with MAtt, GBWAtt requires slightly more SVD-based matrix functions, which leads to a moderately longer training time. To provide a practical efficiency comparison, \cref{tab:training_time} reports the average training time per epoch for each subject on different datasets.

\begin{table}[t]
\centering
\caption{Comparison of average training time (s/epoch) per subject on different datasets.}
\label{tab:training_time}
\resizebox{0.8\linewidth}{!}{
\begin{tabular}{lccc}
\toprule
Model & BCIC-IV-2a & BCI-ERN & MAMEM-SSVEP-II \\
\midrule
MAtt   & 1.84 & 0.98 & 2.55 \\
GBWAtt & 2.47 & 1.43 & 3.32 \\
\bottomrule
\end{tabular}}
\end{table}

\section{Additional Model Interpretation for EEG}
\label{app:add_interp}

For the MAMEM‑SSVEP‑II dataset, \cref{fig:mamem_heat} illustrates gradient responses of the S11 model across channels and time. Dark red regions in the heatmaps indicate strong gradient responses derived from the 8-channel SSVEP EEG recordings. As shown in \cref{fig:mamem_topo}, brain topographic maps highlight robust activations in the visual cortex, particularly at the Oz electrode. Notably, Oz exhibits the most pronounced gradient responses between 0.25 and 0.60 seconds, underscoring its pivotal role in visual processing-consistent with prior studies identifying Oz as the primary SSVEP-responsive channel~\cite{han2018highly,mamem-a1}, further confirming Oz’s central role in SSVEP-based EEG analysis and supporting the validity of our gradient-based interpretation.

For the BCI‑ERN dataset, \cref{fig:ern} shows that gradient responses distinguishing correct from error trials are predominantly centered at electrode FCz. This finding is consistent with extensive empirical evidence implicating the anterior cingulate cortex, located in the central medial prefrontal structure and closely connected with both limbic and frontal areas, plays a key role in generating ERN. Notably, gradient activations at FCz are consistently observed for both feedback types between approximately 0.1 and 0.4 seconds post-stimulus, mirroring known differences in ERP waveforms between correct and incorrect events. These results strongly corroborate the differences in ERP waveforms between correct and incorrect stimuli reported by~\cite{bci-al}.

\end{document}